\documentclass[10pt,twocolumn,letterpaper]{article}

\usepackage{cvpr}              
\usepackage{graphicx}
\usepackage{amsmath}
\usepackage{amssymb}
\usepackage{booktabs}
\usepackage[pagebackref,breaklinks,colorlinks]{hyperref}

\usepackage[capitalize]{cleveref}
\crefname{section}{Sec.}{Secs.}
\Crefname{section}{Section}{Sections}
\Crefname{table}{Table}{Tables}
\crefname{table}{Tab.}{Tabs.}

\def\cvprPaperID{2638} 
\def\confName{CVPR}
\def\confYear{2022}

\begin{document}

\title{SharpContour: A Contour-based Boundary Refinement Approach \\ for Efficient and Accurate Instance Segmentation}

    \author{Chenming Zhu$^{1\dag}$, Xuanye Zhang$^{2\dag}$, Yanran Li$^{4}$, Liangdong Qiu$^{1,2}$, Kai Han$^{5}$, Xiaoguang Han$^{1,3\ddag}$\\
$^1$SSE, CUHK-Shenzhen\ \ \ $^2$Shenzhen Research Institute of Big Data \ \ \ $^3$FNii, CUHK-Shenzhen\ \\$^4$Birmingham University \ \ \ $^5$The University of Hong Kong}

\maketitle

\begin{abstract}
Excellent performance has been achieved on instance segmentation but the quality on the boundary area remains unsatisfactory, which leads to a rising attention on boundary refinement.
For practical use, an ideal post-processing refinement scheme are required to be accurate, generic and efficient. However, most of existing approaches propose pixel-wise refinement, which either introduce a massive computation cost or design specifically for different backbone models.
Contour-based models are efficient and generic to be incorporated with any existing segmentation methods, but they often generate over-smoothed contour and tend to fail on corner areas. In this paper, we propose an efficient contour-based boundary refinement approach, named SharpContour, to tackle the segmentation of boundary area. We design a novel contour evolution process together with an Instance-aware Point Classifier. Our method deforms the contour iteratively by updating offsets in a discrete manner. Differing from existing contour evolution methods, SharpContour estimates each offset more independently so that it predicts much sharper and accurate contours. Notably, our method is generic to seamlessly work with diverse existing models with a small computational cost. Experiments show that SharpContour achieves competitive gains whilst preserving high efficiency.
\end{abstract}

\section{Introduction}

\renewcommand{\thefootnote}{\fnsymbol{footnote}}
\footnotetext[2]{The two authors contribute equally to this paper.}
\footnotetext[3]{Corresponding author: hanxiaoguang@cuhk.edu.cn}

Instance segmentation is a fundamental topic in computer vision which plays important role in scene understanding~\cite{zhao2018psanet}, intelligent robots~\cite{wan2020planning}, clinical analysis~\cite{chen2019learning, zhu2019ace} and autonomous driving~\cite{feng2020deep, hofmarcher2019visual}.
The mainstream of instance segmentation approaches follow the design of detection-then-segmentation framework like Mask R-CNN~\cite{he2017mask} and achieve promising performance. However, high fidelity segmentation of fine details especially in the boundary area remains extremely challenging.

\begin{figure}[t]\begin{center}
   \includegraphics[width=1\linewidth]{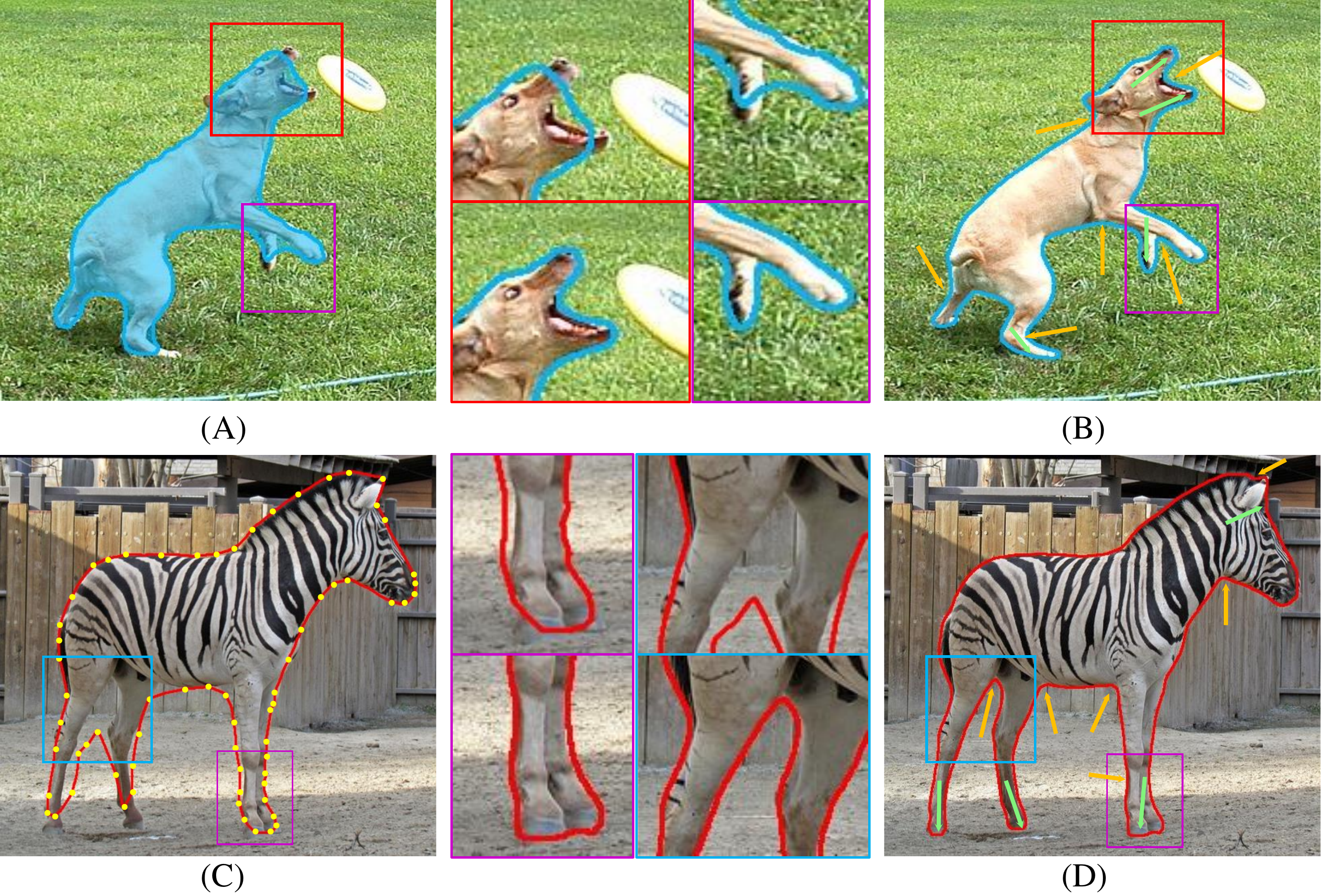}
\end{center}
    \vspace{-12pt}
   \caption{\textbf{Instance segmentation with SharpContour.}  
   \textbf{Top}: A is the coarse mask predicted by Mask R-CNN and B is the refinement result of SharpContour. 
   \textbf{Bottom}: C is the coarse contour generated by DANCE and D is the refinement result of SharpContour. 
   In the corner areas, SharpContour yields significant improvements. 
}
\label{fig:teaser}
\vspace{-15pt}
\end{figure}

To address this issue, the refinement of boundaries is raised as a new topic in recent years. A novel evaluation metric called Boundary AP~\cite{cheng2021boundary} and a number of approaches~\cite{tang2021look, zhang2021refinemask, cheng2020boundary,kirillov2020pointrend} are proposed. Boundary AP~\cite{cheng2021boundary} focuses on the accuracy of the object boundary region thus can better reflect boundary refinement. As for the approaches, refinement can be generally recognized as a post-processing operation which is expected to meet three basic requirements: accurate, efficient and generic. It is worth mentioning that ``generic'' is also a significant factor. However, existing top-performing refinement methods mainly focus on accuracy and lacks consideration of other two factors which are actually important in the practical application scenarios. For example, Boundary Patch Refinement method~\cite{tang2021look} proposed to additionally process patches along boundaries after the instance segmentation model, resulting in massive computational cost. RefineMask ~\cite{zhang2021refinemask} designed a new architecture built upon Mask-RCNN~\cite{he2017mask} to refine the quality of instance mask with fine-grained features. Though the computational cost is not extremely expensive, this method is too specific to be applied to other types of instance segmentation models. It is intrinsically hard for the pixel-wise refinement methods to achieve the three requirements at once, since the dense pixel maps usually bring excessive computation cost.  

Another group of trending segmentation approaches proposed contour-based segmentation methods~\cite{xu2019explicit, peng2020deep, xie2020polarmask, ling2019fast, marcos2018learning, castrejon2017annotating, acuna2018efficient} which directly process and generate sparse points along the boundaries. They are advantageous on efficiency and naturally generic to be appended after mask-based algorithms since it is straightforward to extract a contour from a mask. Thus, an interesting question can be raised up: `` Is it possible to use contour-based methods to address the boundary refinement problem?" 

However, existing contour-based methods, like DANCE~\cite{liu2021dance} and DeepSnake~\cite{feng2020deep}, tend to produce over-smoothed contours especially on the sharp turner areas (as shown in Fig.~\ref{fig:teaser}) so that their performances on boundary areas is still lagging behind. We figure out the major reason which leads to the oversmoothing problem is that all the vertices on the contour are tied together by their feature learning strategies (e.g. circular convolutions~\cite{feng2020deep}) and the smoothness regularization in their contour evolution process. As a result, a slight offset on one vertex will cause a wide chain reaction on all vertices of the contour. To address this, we elaborate a drastically different contour evolution process that estimates the deformation offsets independently for each vertex. Based on this, we propose an \textit{accurate, efficient and generic} refinement approach, named \textit{SharpContour}, for boundary area using contour-based representation. 

Specifically, our SharpContour takes a coarse contour as input and deforms each vertex on the contour individually. To avoid artefacts caused by the involved ``independency'', we constrain each vertex to move along its normal direction and conduct the deformation in an iterative procedure, making the deformation stably. For the sake of  balancing between efficiency and accuracy for each vertex's movement, instead of directly regressing the offset, we propose to determine the target position by performing classification on a few discrete points. Specifically, we first sample some points along the moving direction and classify them into ``inner/outer'' status to obtain the flipping position which indicates the boundary. 
The classifier plays a pivotal role in this contour evolution process. Thus, we carefully design an Instance-aware Point Classifier (IPC), which can predict ``inner/outer'' of a vertex with respect to different instance boundaries in the form of a probability score. There are two important factors for IPC to achieve high fidelity: 1) It is required to be instance-aware, which is essential to enable generating different results for a pixel with respect to different instances. For this purpose, the parameters of the IPC are predicted on-the-fly for each instance. 2) It should capture the information of the boundary details. Thus IPC takes the fine-grained feature, which is derived from the high-resolution feature map, as input to predict the vertex state.

With IPC, we can determine how to move each vertex during the deformation. The moving distance is further defined by the object size and IPC probability score. We iteratively adjust the vertices of the predicted contour, until they match the object boundary. In this process, SharpContour avoids the high-dependency over neighbouring vertices without using any smooth term, thus overcoming the oversmoothing issue. Notably, SharpContour only introduces a handful of parameters and involves a negligible amount of points in the calculation. Therefore, our approach is considerably efficient. Experiments manifest that SharpContour can work seamlessly with various segmentation models by introducing a small computational cost and producing high-quality object boundaries.

We validate the effectiveness of our SharpContour approach quantitatively and qualitatively on various large scale segmentation benchmarks. On COCO datasets~\cite{lin2014microsoft}, we use both AP and the Boundary AP metrics. Coupled with DANCE~\cite{liu2021dance}, SharpContour brings significant improvements of \textbf{1.5 AP} and \textbf{3.2 Boundary AP}. Coupled with the Mask R-CNN~\cite{he2017mask} and CondInst~\cite{tian2020conditional} as the refinement approach, our approach surpasses the baseline by a significant gain of \textbf{2.3, 2.2 AP} and \textbf{3.4, 3.3 Boundary AP} respectively. On Cityscapes datasets~\cite{cordts2016cityscapes}, SharpContour achieves \textbf{3.9 AP} boost when combined with Mask R-CNN~\cite{he2017mask}. Compared with other boundary refinement models, SharpContour consistently produces high-quality boundaries with a low computational cost. Taking the output of the state-of-the-art contour refinement method, RefineMask, as initial contour, SharpContour still gains \textbf{0.5 AP} and \textbf{1.1 Boundary AP}. The qualitative results further demonstrate the effectiveness of our approach.

\begin{figure*}[!ht]
\begin{center}
   \includegraphics[width=0.88\linewidth]{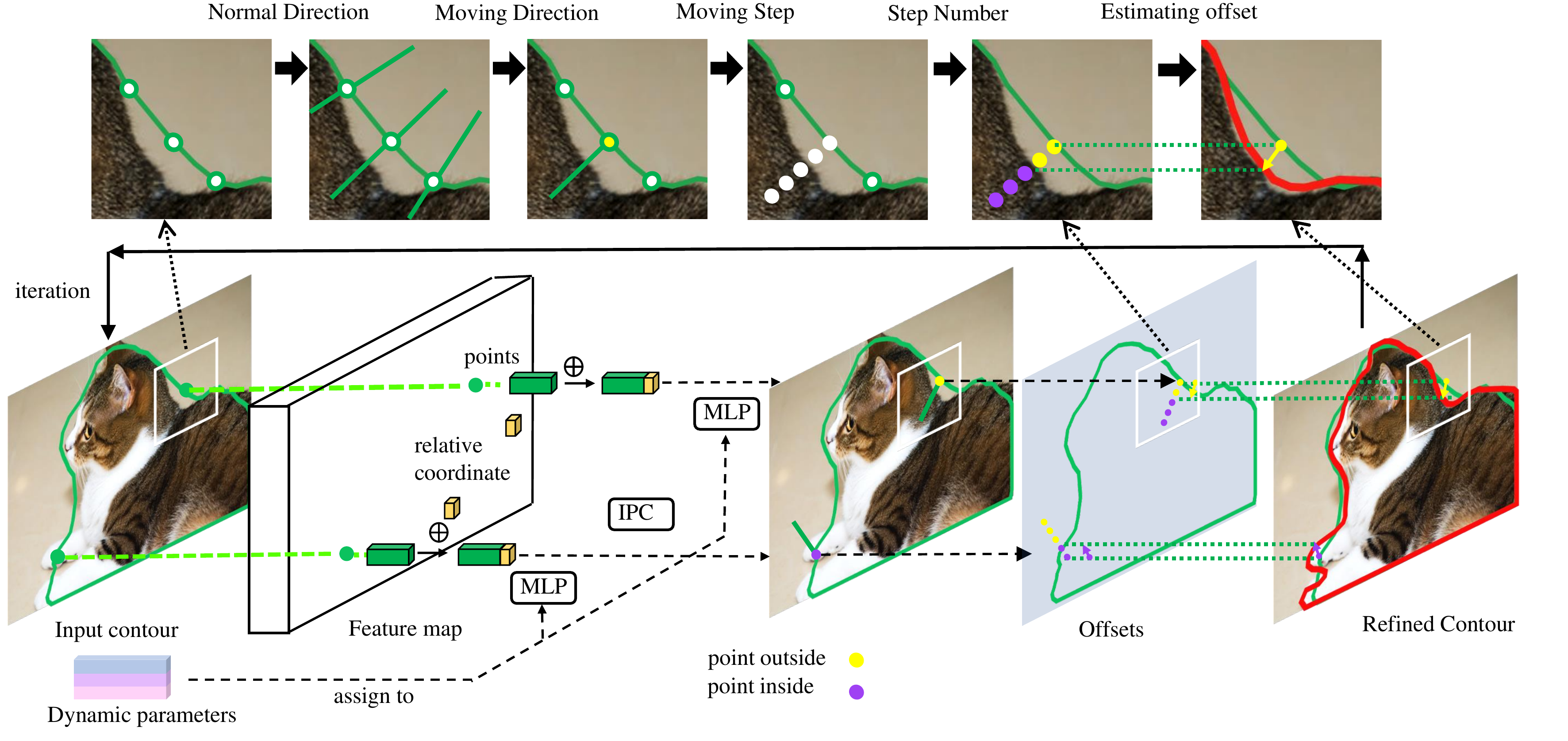}
\end{center}
    \vspace{-20pt}
   \caption{\textbf{Contour Evolution of SharpContour.} SharpContour obtains the initial contour from coarse segmentation results and deforms the contour to achieve boundary refinement. In the deformation process, 1) SharpContour obtains the normal direction. 2) SharpContour predicts the inner/outer state of vertex to decide negative/positive normal direction of deformation. 3) SharpContour decides the step size. 4) SharpContour obtains the moving step number based upon the position of the \emph{flipping point}. 5) SharpContour estimates the offsets and deforms the contour.
}
\label{fig:Method:Pipeline}
\vspace{-18pt}
\end{figure*}

\section{Related Work}
\textbf{Mask-based methods.}
The mainstream mask-based approaches can be broadly categorized into top-down and bottom-up methods. The typical top-down approaches~\cite{he2017mask, liu2018path, chen2019hybrid, cheng2020boundary, bolya2019yolact, wang2020solo, chen2020blendmask, lee2020centermask, zhang2020mask, tian2020conditional} follow the pipeline of Mask-RCNN~\cite{he2017mask}, which detects each instance first and predicts a pixel-wise binary map for each instance. Models can be categorized into two-stage~\cite{liu2018path, chen2019hybrid, cheng2020boundary} and one-stage~\cite{bolya2019yolact, wang2020solo, chen2020blendmask, lee2020centermask, ying2019embedmask, tian2020conditional} according to their pipelines. However, they usually rely on a pooling operation to extract canonical-size features from the feature maps for each instance, which loses many boundaries details. CondINST~\cite{tian2020conditional} removed the RoI operation and use dynamic convolution layer to generate segmentation masks, but it still losses many details around boundaries. Bottom-up approaches~\cite{zhang2015monocular,  newell2016associative, kirillov2017instancecut, liu2018affinity, sofiiuk2019adaptis, gao2019ssap} first generate semantic segmentation and then cluster pixels into different instances. In all the pixel-wised methods, the boundary points only occupy a very small proportion (less than 1\% \cite{kirillov2017instancecut}) so that their accuracy is usually over compromised during the optimization. Our method avoids these drawbacks and can work together with various models to enhance their performance on boundaries.

\textbf{Boundary refinement}
To tackle the aforementioned limitations, various boundary refinement methods~\cite{kirillov2020pointrend, zhang2021refinemask, tang2021look, liang2020polytransform,cheng2020boundary, yuan2020segfix} are proposed. Meanwhile, to evaluate the boundary performance more accurate, a novel evaluation metric Boundary AP~\cite{cheng2021boundary} is proposed, which can solve the desensitization problem of the current evaluation metric (AP) for object boundary region. As for the approaches, BPR~\cite{tang2021look} adopts a post-processing scheme, refining predicted instance boundary patches in detail to improve mask quality. Another top-performing model, PolyTransform ~\cite{liang2020polytransform} proposes the first contour-based refinement approach which uses the results of the mask-based model as initial contour and refines the contour by Transformer network~\cite{vaswani2017attention}. These methods achieve superior performance while introducing a large computational cost and scarifying the efficiency. Some other methods try to balance the performance enhancement and the extra computational overheads. For example,  PointRend~\cite{kirillov2020pointrend} only performs point-based segmentation around the boundary blurred areas to obtain high-quality mask. RefineMask~\cite{zhang2021refinemask} achieves high-quality results by introducing fine-grained features during the upsampling process to refine entire objects. 
However, these two methods lack of generality since they can only refine the results of specific mask-based methods. In contrast, our refinement approach can not only achieve accurate and efficient refinement but also be generic to refine the results of both mask-based and contour-based models. 

\textbf{Contour-based methods}
Neural contour-based models~\cite{rupprecht2016deep, marcos2018learning, xu2019explicit, jetley2017straight, ling2019fast, xie2020polarmask} for segmentation are widely studied in recent years due to their potential advantage on efficiency. Most of them formulate the contour as polygons and propose various regression methods to estimate the coordinate of vertices on the polygons.~\cite{marcos2018learning, xu2019explicit} propose CNN-based models which can learn high-level and expressive features for deformation.~\cite{ling2019fast} proposes a contour-based model which utilizes Graph Convolution Networks (GCN) to regress the offsets of vertices on the contour. Following the pipeline of traditional Snake model~\cite{kass1988snakes}, DeepSnake~\cite{peng2020deep} proposes a two-stage contour evolution process and designs circular convolution to exploit the features on the contour. DANCE~\cite{liu2021dance} follows the pipeline of DeepSnake~\cite{peng2020deep} and introduces edge attention module and improves the matching scheme in the contour evolution process. DANCE~\cite{liu2021dance} achieves state-of-the-art results for the contour-based instance segmentation methods. However, all of these regression-based contour evolution methods still suffer severe performance degradation in approximating the corners or cusps of instance. This is because their regression method needs to balance the regression error and the smoothness of the current contour and the offsets are highly dependent on each other during feature learning. Our approach elaborates a drastically different evolution method. In our deformation stage, the offset of each vertex on the polygon is estimated by discrete moving steps rather than regression. Incorporated with DANCE~\cite{liu2021dance}, our approach is proved to be more effective than all the existing contour-based methods, yielding accurate and high-fidelity segmentation results even for significant challenging areas.

\section{Methodology}
We propose a contour refinement approach, SharpContour, which coupled with existing instance segmentation models (either mask-based or contour-based), can produce high-quality boundary segmentation results.
Let $C^{(0)} : \{ x_i \ | i = 1,\ ...,\ N \}$ be an initial contour, which is  obtained by off-the-shelf instance segmentation methods. The contour is defined by a sequence of vertices $x_i$.
SharpContour iteratively moves the vertices to approach the actual boundary of an instance. 
To perform the contour evolution, we design an iteration process to deform the vertices along their normal. The key challenge of the contour evolution methods is to predict the offset for each vertex accurately. In contrast to existing methods that directly regress the offsets for all the vertices at once, which is often error-prone and leads to over-smoothed contours, we propose to adjust the vertices iteratively in a discrete manner inspired by~\cite{sharf2006competing}.
Due to the inherent challenge of directly regressing offsets, we cast this problem to be a classification-based formula.
Specifically, we propose an Instance-aware Point Classifier (IPC) $\varphi(x_i)$ which predicts a state for the vertex $x_i$ indicating its relative position to the actual object boundary (i.e., inside or outside), such that we can determine whether a vertex should march along the positive or negative normal direction. 
Our approach greatly reduces the dependency on the offsets of neighbour vertices and allows for delicate segmentation details on complex and challenging boundary areas, such as the corner and cusps areas where the existing methods struggle. An overview of SharpContour can be found in~\cref{fig:Method:Pipeline}.

\subsection{Contour Evolution}
\label{sec: Deformation}
Given a vertex $x_i$ which is off the object boundary, its evolution process in one iteration can be written as  
\begin{equation}
    x'_i = x_i + ms_id_i,
\end{equation}
where $d_i$ is the moving direction, $s_i$ the moving step and $m$ the step number.
For each vertex, $\varphi(x_i)$ will output a scalar value in $[0, 1]$, indicating the probability of $x_i$ being outside (i.e., $\varphi(x_i)$ = 1) or inside (i.e., $\varphi(x_i)$ = 0) the object. $\varphi(x_i)$ = 0.5 means that IPC is not sure whether the point is inside or outside the object, implying the point is probably located on the boundary. More details of IPC will be introduced in~\cref{sec: IPC}.
Next, we describe how we obtain the moving step $s_i$  and the step number $m$. They together define the moving distance. 

To define the moving step, we take into account the object size (in terms of the area of the bounding box, denoted as $A$) and the uncertainty of the vertex state from IPC. 
Intuitively, if the object is larger, the step size should also be larger, and vice versa. If the uncertainty of vertex is higher, the step size should be smaller to reach finer prediction, and vice versa. To reflect these, concretely, we define the step size as $s_i = \lambda \sqrt{A} |\varphi(x_i)-0.5|$, where $\lambda$ is an empirical deformation ratio which we set to $0.003$ in experiments. $|\varphi(x_i)-0.5|$ indicates the uncertainty of state of $x_i$ for IPC, the closer $\varphi(x_i)$ is to 0.5, the higher the uncertainty of state of $x_{i}$ for IPC, so smaller step size for $x_i$ should be assigned for accuracy.  
For the moving step number, ideally, we expect the vertex to progressively move towards the actual object boundary. Therefore, we move the vertex along $d_i$ step-by-step, and after each move, we examine the value of $\varphi$ on the current location. If we reach a location that $\varphi$ indicates the vertex is moved from inside to outside of the boundary or the other way around, then we stop. We call this location \emph{flipping point}. The number of moves from the original location to the flipping point is then our moving step number $m$. To avoid the vertex moving towards an improper direction too far away, we set an upper bound for the move, i.e., for each move if after $M$ steps, a flipping point is still not reached, we then set $m=M$. 
\paragraph{Iteratively Evolution} 
We perform the above evolution process for each vertex of $C^{(0)}$ to obtain an updated contour $C^{(1)}$. We then run another evolution cycle for $C^{(1)}$ to update the vertices. We iteratively run the evolution process and generate 
a series of contours $\{ C^{(0)},\ C^{(1)},..., \ C^{(n)}, ...\}$ to approach actual object boundary progressively. 
Note that during each iteration, if a vertex reaches a \emph{flipping point}, which indicates that it has reaches the object boundary, it will not participate the subsequent deformation process anymore. This further improves the efficiency of SharpContour. 

\begin{figure}[t]
\begin{center}
  \includegraphics[width=0.7\linewidth]{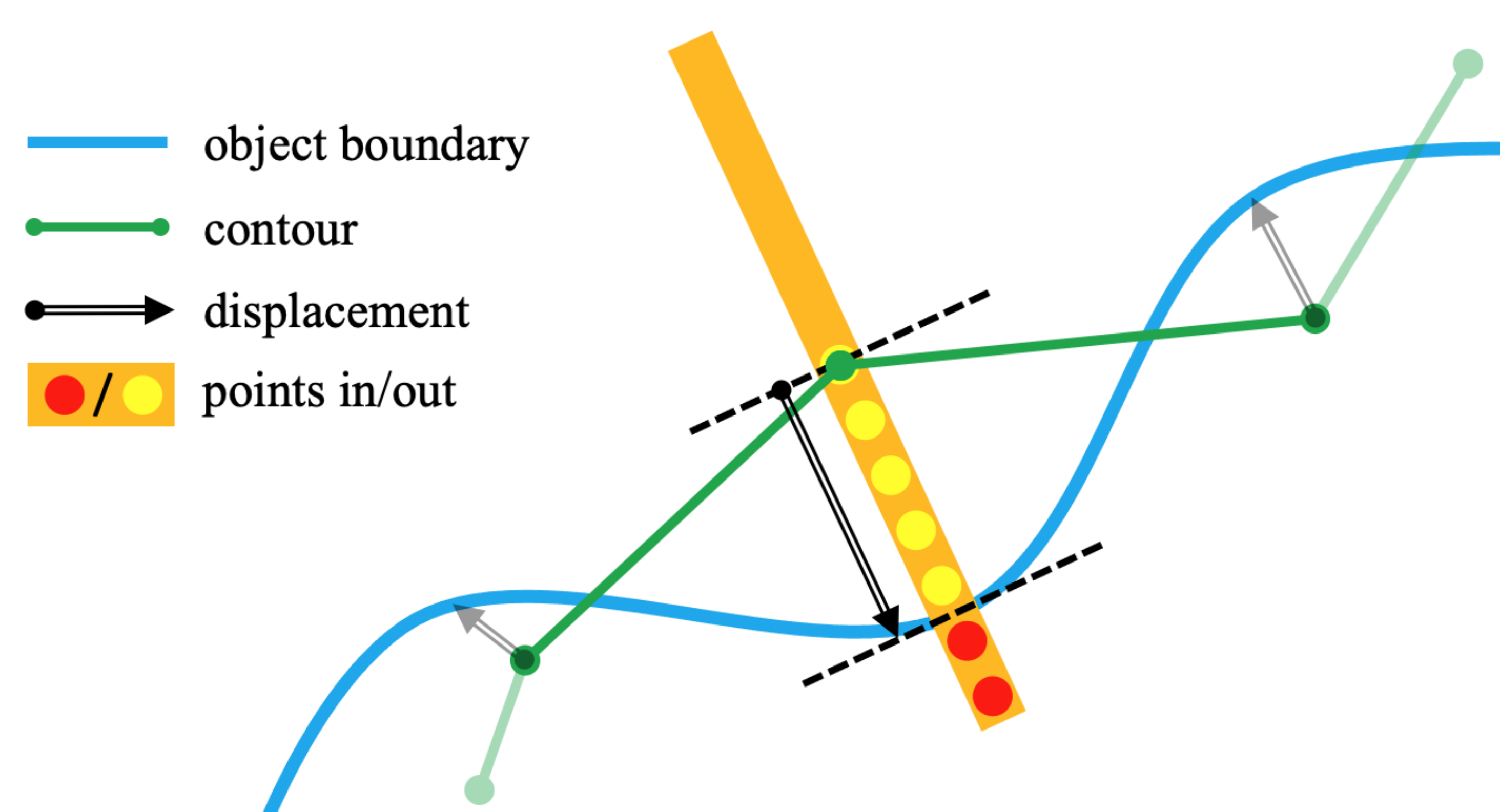}
\end{center}
\vspace{-18pt}
  \caption{\textbf{Deformation of SharpContour.} Value flipping indicates the position of instance boundary. SharpContour uses the position of flipping points to update the contour. 
}
\label{fig:Met:Def}
\vspace{-15pt}
\end{figure}

\subsection{Instance-aware Point Classifier} 
\label{sec: IPC}
Here, we introduce our Instance-aware Point Classifier (IPC) $\varphi$, which aims at predicting a state for a given vertex $x_i$ to tell its relative position to the actual object boundary. In a nutshell, IPC takes the fine-grained feature of $x_i$ derived from the high-resolution feature map and the relative location of $x_i$ to its corresponding bounding box as input to predict the probability indicating whether $x_i$ is located inside or outside the object. Importantly, the parameters of $\varphi$ are dynamically predicted for each instance on the fly, such that $\varphi$ is instance-aware, which plays an important role at inferring vertices relative location to the instance.
\paragraph{Fine-grained Feature}
To be able to identify the subtle details for the vertex evolution, we employ the fine-grained features from the high-resolution feature map produced by the instance segmentation backbone as the input for $\varphi$.
We append a convolution layer on the high-resolution feature map to further encode feature to benefit $\varphi$ for contour refinement and reduce the dimension. Let the encoded fine-grained feature for $x_i$ be $f_i$.we concatenate $f_i$ with the relative coordinates $c_i$ of $x_i$ to the instance bounding box, forming a new fine-grained location-aware feature vector $f_i^c = [f_i; \ c_i]$, which serves as the input of $\varphi$ to predict the state probability.
Taking Mask R-CNN as example, we utilize the highest resolution feature map from the feature pyramid network (FPN), which is $1/4$ of the image size. To decrease the parameters number of IPC and enhance the utilization of the high-resolution feature, we attach a $3\times3$ convolution layer to this feature map and reduce its channel from $256$ to $16$. From the resulting feature map, we extract the fine-grained feature for each vertex.

\begin{figure*}[tb]
\begin{center}
   \includegraphics[width=0.95\linewidth]{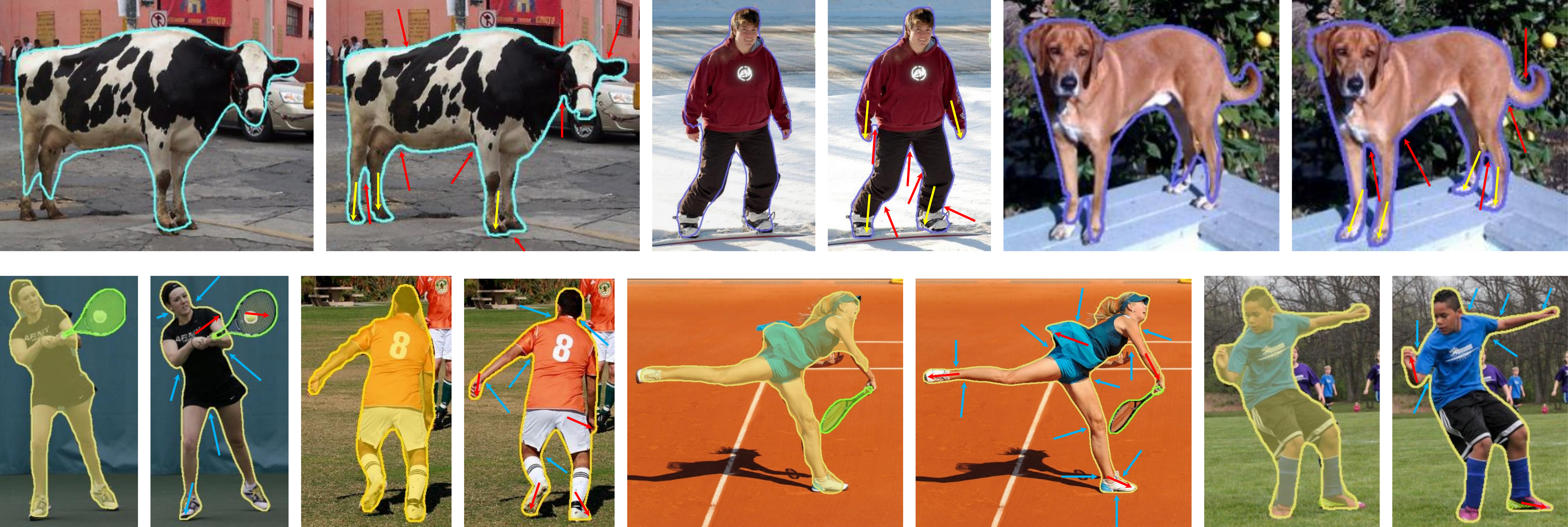}
\end{center}
\vspace{-18pt}
   \caption{\textbf{Qualitative results on COCO dataset.}  We use SharpContour to refine the segmentation results of different models. The top line is the results of DANCE while the bottom line is the results of Mask-RCNN. For each example, the left is the result before refinement while the right is our result. As is illustrated, SharpContour can refine the segmentation results near instance boundary. 
}
\label{fig:Exp:City}
\vspace{-10pt}
\end{figure*}

\paragraph{Instance-aware Dynamic Parameters}
Only equipped with the fine-grained feature, the classifier still struggles at the determination. For example, a point with the same feature can be outside one instance but inside another instance. To deal with this problem, our classifier needs to be instance-aware and has strong ability to grasp holistic information to identify each instance. Inspired by \cite{tian2020conditional}, we propose to dynamically \emph{predict} the parameters $\theta$ of the classifier $\varphi$ for each instance, based on the feature of each instance. This strategy can capture strong instance-aware information therefore it can discriminate the state of a vertex w.r.t different instance. For example, we incorporate a boundary controller head inspired by ~\cite{tian2020conditional} for Mask R-CNN to obtain the dynamic parameters for each instance. The boundary controller head is a very compact and light-weight network with three fully connected layers. 
The output dimension of this controller head is the same with the parameter number of IPC.
By sharing feature with the original mask head of Mask R-CNN, we can make better use the instance-aware feature while only introducing a small computation cost.
Our IPC can therefore be written as $\varphi(x_i) = \varphi_{\theta}([f_i, c_i])$. If $\varphi(x_i) > 0.5$, $x_i$ is considered as outside the object predicting the label $\hat{y}_i$ as $1$, otherwise inside the object predicting the label $\hat{y}_i$ as $0$.
In our implementation, IPC is simply realized an MLP with three hidden layers with the ReLU activation. The sigmoid activation applied to its output layer. Hence, the extra cost introduced by the IPC is marginal.

\subsection{Instance Segmentation with SharpContour} 
\label{sec:workinmodels}
In this section, we describe how to combine our SharpContour with different types of instance segmentation models, including mask-based models and contour-based models. In the training stage, we train the head which generates parameters of IPC by randomly sampling pixels near the ground-truth instance boundary. In the inference stage, SharpContour obtains the initial contour from the coarse segmentation results and the contour evolution process can be iterated for refinement. 
In general, when working with existing instance segmentation methods, the SharpContour takes three inputs: initial contour, fine-grained features and dynamic parameters. Methods to obtain these three inputs vary from different instance segmentation methods. Here we introduce how to apply SharpContour to three typical instance segmentation methods. 
For contour-based models such as DANCE~\cite{liu2021dance}, the predicted contour of their model can be used directly as the initial contour for our SharpContour. For Mask-based models such as Mask R-CNN~\cite{he2017mask} and CondInst~\cite{tian2020conditional}, we extract the initial contour by transferring the predicted mask into contour representation. 
To obtain the fine-grained features, we use convolution layers to encode the feature maps of highest resolution from their backbone and use the encoded feature map to extract fine-grained features.  
To obtain the dynamic parameters, we tailor the boundary controller head to incorporate it with various instance segmentation models. The detailed network structure and parameters settings are provided in the supplementary. 

\subsection{Loss Function}
\label{sec:loss}
We use the focal loss~\cite{lin2017focal} to train the boundary controller head which generates parameters of IPC and the convolution layer which encodes the fine-grained features. Since the points sampled around ground-truth boundary for training is random, the ratio between positive and negative samples is not fixed. To compensate this, we exploit the focal loss with a dynamic coefficient $\alpha$, which is the ratio of current positive and negative samples. The loss can be written as
\vspace{-10pt}
\begin{equation}
    L_{IPC} = \left\{\begin{aligned}
                    -\alpha(1-\hat{y}_i)^{\gamma}log(\hat{y}_i),\ & y_i=1 \\
                    -(1-\alpha)(\hat{y}_i)^{\gamma}log(1-\hat{y}_i), \ & y_i=0 \\
                            \end{aligned}\right.
\label{eq:dy_Foc}
\vspace{-5pt}
\end{equation}
where $\hat{y}$ denotes the predicted label, $\alpha$ is dynamically determined by the proportion of positive and negative samples in the current batch and $\gamma$ is the difficulty factor which we set to 2 in our experiments. 

After combining our SharpContour with a baseline instance segmentation  model, we jointly train our SharpContour and with the coupled instance segmentation model.  The overall loss function can be expressed as
\vspace{-8pt}
\begin{equation}
L = L_s + \mu L_{IPC},
\vspace{-8pt}
\end{equation}
where $L_s$ is the original loss of the instance segmentation model. We set the weight $\mu=10$ in our experiments.

\begin{table*}[tb]
\centering
\resizebox{0.82\textwidth}{!}{
\begin{tabular}{c|c|c|c|c|c|c|c|c|c}
\hline
        & $AP_{dev}$ & $AP_{val}$ & Boundary AP & AP* & $AP*_{S}$ & $AP*_{M}$ & $AP*_{L}$ & FPS & Inference time (ms/per image) \\ \hline
Mask R-CNN                              & 34.6 & 34.7 & - & 36.8 & 22.6 & 43.7 &52.0 & 12.3 &  81.3      \\
Mask R-CNN*                             & 35.5 & 35.2 & 21.2 & 37.6 & 22.8 & 44.7 & 53.8 & 17.5 & 57.1        \\
DANCE                                   & 34.6 & 34.5 & 20.2 & - & - & - & - & 16.5 &  60.6   \\ 
CondInst                                & 35.4 & 35.7 & 21.6 & - & - & - & - & 17.4 &  57.5      \\ \hline
PointRend                               & 36.8 & 36.3 (+1.1)& 23.5(+2.3) & 39.7& 22.9 & 46.7 & 57.4 & 13.0 & 76.9 (+19.8) \\
RefineMask                              & 37.6 & 37.3 (+2.1)& 24.7(+3.5) & 40.9& 24.1 & 48.8 & 58.0 & 13.0 & 76.9 (+19.8)\\ \hline
DANCE + SharpContour                    & 36.3 & 36.1 (+1.5)& 23.9(+3.7) & -  & - & - & - & 13.6 & 73.5 \\ 
Mask R-CNN* + SharpContour              & 37.7 & 37.5 (+2.3)& 24.6(+3.4) & 41.2& 24.2 & 49.1 & 58.5 & 15.0 & 66.7 (+9.6)\\
CondInst + SharpContour                 & 37.7 & \textbf{37.9} (+2.2)& 24.9(+3.3) & -  & - & - & - & 15.4 &  64.9  \\ \hline
RefineMask + SharpContour               & \textbf{38.0} & 37.8 (+0.5)& \textbf{25.8(+1.1)} & \textbf{41.9} & \textbf{24.3} & \textbf{49.4} & \textbf{59.1} & 12.1 &  82.6 (+5.7)\\ \hline
\end{tabular}}
\centering
\vspace{-8pt}
\caption{\textbf{Comparisons on COCO val2017 and test-dev.}
AP$_{dev}$ denotes the evaluation results on $test-dev$, and other
columns denotes the evaluation results on $val2017$. ``Mask R-CNN" is a original Mask R-CNN, and ``Mask R-CNN*" is the improved version in Detectron2. All methods are trained with 1x schedule using R50-FPN backbone. The FPS is measured on a single Tesla V100 GPU. SharpContour brings significant AP enhancement for DANCE, Mask R-CNN and CondInst. Moreover, SharpContour can achieve competitive performance compared with other boundary refinement approaches with the highest efficiency.}
\label{tab:Comp:COCO}
\vspace{-23pt}
\end{table*}
\begin{table}[ht]
\centering
\vspace{5pt}
\resizebox{0.4\textwidth}{!}{
\begin{tabular}{c|c|c|c|c}
\hline
Method      &  AP  & AP$_{S}$  & AP$_{M}$ &AP$_{L}$  \\ \hline
Mask R-CNN  &  33.8 & 12.0 & 31.5 & 51.8     \\
PointRend   &  35.8(+2.0) & - & - & - \\
RefineMask  &  37.6(+3.8) & \textbf{14.6} & 34.0 & 58.1 \\\hline
Mask R-CNN + SharpContour   & \textbf{37.7(+3.9)} & 14.4 & \textbf{34.2} & \textbf{58.3}     \\ \hline
\end{tabular}
}
\centering
\vspace{-8pt}
\caption{\textbf{Results on Cityscapes validation set.} The training setting for all models are same: trained on fine annotations for 64 epochs, using multi-scale training and ResNet-50 with FPN.}
\label{tab:comp:Cityscapes}
\vspace{-18pt}
\end{table}

\section{Experiments}
\subsection{Impementation Details}

\paragraph{Training strategy}
We train the SharpContour with the DANCE, Mask R-CNN and the CondInst together to enhance the performance. For each combination model, we conduct experiments using totally the same training settings with coupled model, including training epoch, learning rate schedule, data augmentation methods, etc, to ensure fairness. We set the deformation ratio $\lambda =0.003$, the number of sampling points $M=10$ and the resolution of the polygon $N=128$. The iteration number of evolution is set to be 3. More detailed training settings will be described in \textit{supp}.

\subsection{Benchmarks and Metrics}
We evaluate the efficiency and the effectiveness of SharpContour on three standard benchmarks, which are Cityscapes~\cite{cordts2016cityscapes} and Microsoft COCO~\cite{lin2014microsoft}. Following the prior work, we use Mask AP as the evaluation metrics. To further demonstrate the effectiveness of SharpContour, we report Boundary AP on the COCO datasets. (Unless explicitly stated, AP denotes Mask AP in this paper. )

\noindent \textbf{COCO}~\cite{lin2014microsoft} is one of the most common benchmarks to evaluate the models for object detection and segmentation tasks, and we mainly report the results on COCO. Our models were trained on train2017. Following PointRend and RefineMask, we also report AP$^{*}$, which evaluates the COCO categories using LVIS annotations, since LVIS annotations have much higher quality mask. The results of AP$^{*}$ are obtained by the same model trained on COCO.

\noindent \textbf{Cityscapes}~\cite{cordts2016cityscapes} is a large-scale dataset for pixel-level and instance-level segmentation evaluation, which provides a massive amount of video records of daytime urban street scenes from $50$ cities. Cityscapes contain a rich set of annotations including $5,000$ images with fine pixel-level annotations and $20,000$ images with coarse annotations. It is one of the most widely used and challenging benchmarks.

\subsection{Comparison with the state-of-the-art}
\noindent \textbf{Effectiveness for contour-based model}
We apply SharpContour to the DANCE~\cite{liu2021dance}, which follows the idea of DeepSnake~\cite{peng2020deep} and achieves current state-of-the-art performance for contour-based instance segmentation methods. We conduct experiments on the COCO datasets. As is shown in the Tab.~\ref{tab:Comp:COCO}, SharpContour brings 1.5AP and 3.7 Boundary AP improvement.  

\noindent \textbf{Effectiveness for Mask-based methods}
We apply SharpContour to the Mask R-CNN~\cite{he2017mask} and the CondInst~\cite{tian2020conditional} and complete experiments on the COCO and Cityscapes datasets. On COCO datasets, we further report the AP* measured using the higher quality LVIS~\cite{gupta2019lvis} annotations. 1) For Mask R-CNN~\cite{he2017mask}, SharpContour outperforms baseline model 2.3AP, 3.4 Boundary AP and 3.6AP* on COCO datasets (Tab.~\ref{tab:Comp:COCO}) and yields 3.9AP enhancement on CityScapes datasets(Tab.~\ref{tab:comp:Cityscapes}). 2) For CondInst~\cite{tian2020conditional}, SharpContour achieves an improvement of 2.2AP and 3.3 Boundary AP on the COCO~\cite{lin2014microsoft} datasets. Besides, `RefineMask + SharpContour' combination in Tab.~\ref{tab:Comp:COCO} represents: we directly run `Mask R-CNN + SharpContour' model on the contour simply extracted from the output mask of RefineMask, it still gains 0.5AP and 1.1 Boundary AP improvement. SharpContour gains larger improvements on Cityscapes compared with COCO. The probable reason is that the classes of Cityscapes is much less than COCO, which leads to a better IPC. This further proves the effectiveness of SharpContour.

\noindent \textbf{Efficiency}
We compare the efficiency with other boundary refinement methods, including PointRend~\cite{kirillov2020pointrend} and RefineMask~\cite{zhang2021refinemask}, on the COCO and Cityscapes dataset. For fairness, 1) These methods are built on Mask R-CNN. 2) All the methods adopt ResNet-50 with FPN as the backbone, 3) All the models are trained for the same epochs. As is shown in Tab.~\ref{tab:Comp:COCO},~\ref{tab:comp:Cityscapes}, comparing with PointRend~\cite{kirillov2020pointrend}and RefineMask~\cite{zhang2021refinemask}, SharpContour achieves better performance with higher efficiency. Notably, PointRend~\cite{kirillov2020pointrend} and RefineMask~\cite{zhang2021refinemask} incorporate their methods after Mask R-CNN and introduce additional inference time of 19.8ms. Compared with them, our SharpContour only introduce inference time of 9.6ms, half of theirs, but brings more gains.  

\noindent \textbf{Qualitative Comparison} \ Fig.~\ref{fig:Exp:City} shows the qualitative results of SharpContour on the COCO datasets. We use SharpContour to refine two models: DANCE (top line) and Mask R-CNN (bottom line). For each instance in these figures, the left one is the result without the SharpContour while the right one is the result with the SharpContour. As is illustrated, the SharpContour can ameliorate the contour near instance boundary since it can extract extra information in vertex neighbourhood so that achieve high-fidelity and accurate results for diverse instances. More results can be found in the supplementary materials. Fig.\ref{fig:qua_comp} shows qualitative comparison results between RefineMask and SharpContour+MaskRCNN. SharpContour can deal with objects with complex contour and perform better at challenging regions (e.g., thin structure region). 

\subsection{Ablation Study}
We conduct ablation experiments to analyze the proposed SharpContour on the COCO datasets. We explore the influence of different choices of SharpContour. 

\noindent \textbf{Training Scheme}
To demonstrate the effectiveness of SharpContour in advance, we also adopt the training scheme which freezes all parameters of the instance segmentation models and only trains the SharpContour independently. Specifically, we only train the boundary controller head (Sec.~\ref{sec: IPC}) used to generate parameters of the IPC and the convolution layer (Sec.~\ref{sec: IPC}) used to generate fine-grained feature. As shown in Tab.~\ref{tab:Abl:Scheme}, SharpContour can still refine the instance segmentation results of these models by adopting such training scheme. 

\begin{figure}[t]\begin{center}
\vspace{-5pt}
   \includegraphics[width=1\linewidth]{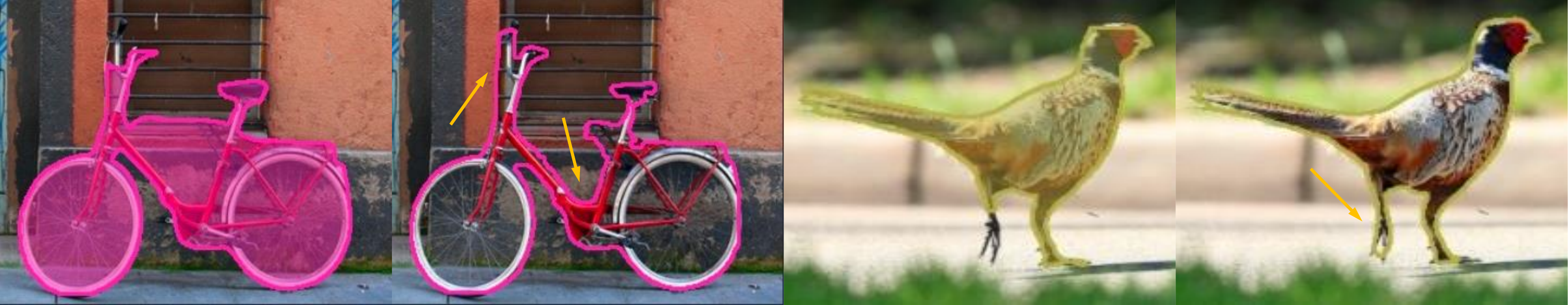}
\end{center}
\vspace{-18pt}
\caption{\textbf{Qualitative comparison results of complex contours} SharpContour+MaskRCNN (Right) performs better than RefineMask(Left) at challenging regions. }
\label{fig:qua_comp}
\vspace{-10pt}
\end{figure}

\noindent \textbf{Effectiveness for feature extraction}
SharpContour shares the backbone with the instance segmentation model to refine. We explore what effects such training strategy casts to the feature extraction process of the backbone. We examine the instance segmentation accuracy of the model w/o training with the SharpContour. As illustrated in Tab.~\ref{tab:Abl:Effects}, Training with SharpContour can actually enhance the performance of the instance segmentation models, which indicates that SharpContour can ameliorate feature extraction process. 

\noindent \textbf{Effects of dynamic step size}
The output probability of the IPC indicates the uncertainty of inner/outer state. We argue that such uncertainty can be used to better control the moving distance in contour evolution process. We conduct ablation experiments w/o such dynamic step size on COCO datasets. As reported in Tab.~\ref{tab:Abl:Adaptive}, the dynamic step size yields 0.5AP and 0.9AP* enhancements. Moreover, we notice that dynamic step size can also stable the inference process. 

\begin{table}[t]
\resizebox{0.3\textwidth}{!}{
\begin{tabular}{c|c|c}
\hline
Training Scheme        & AP       & AP*           \\ \hline
joint training          & 37.5(+2.3)    & 41.2(+3.6)          \\
training alone          & 36.2(+1.0)     & 39.8(+2.2)           \\\hline
\end{tabular}}
\centering
\vspace{-8pt}
\caption{\textbf{Results of Different training scheme on COCO val2017.} The performance is reported on ``Mask R-CNN + SharpContour" combination. }
\vspace{-12pt}
\label{tab:Abl:Scheme}
\end{table}

\begin{table}[t]
\resizebox{0.45\textwidth}{!}{
\begin{tabular}{c|ccc}
\hline
      & Mask R-CNN & Mask R-CNN$^{+}$  & Mask R-CNN+SharpContour \\ \hline
AP    & 35.2       & 35.5            & 37.5                      \\
$AP^*$   & 37.6       & 38.0            & 41.2                      \\ \hline
\end{tabular}}
\vspace{-8pt}
\centering
\caption{\textbf{Results on COCO val2017}. Mask R-CNN$^{+}$ is the results of Mask R-CNN trained with SharpContour. It demonstrates that original model can get benefits from joint training, and our proposed module indeed improve the mask quality. }
\label{tab:Abl:Effects}
\vspace{-18pt}
\end{table}

\begin{table}[t]
\resizebox{0.25\textwidth}{!}{
\begin{tabular}{c|c|c}
\hline
adaptive step        & AP             & AP*           \\ \hline
w                    & 37.5           & 41.2         \\
w/o                  & 37.0(-0.5)     & 40.3(-0.9)          \\\hline
\end{tabular}}
\centering
\vspace{-8pt}
\caption{\textbf{Effectiveness of adaptive step on COCO val2017}. When combining Mask R-CNN with our SharpContour, the adaptive step strategy can bring significant 0.5AP and 0.9AP* improvements, which can also stable the inference process. }
\label{tab:Abl:Adaptive}
\vspace{-12pt}
\end{table}

\noindent \textbf{Larger models, longer training}
We train SharpContour together with Mask-RCNN using different backbones, including ResNet-101 with FPN and ResNeXt-101 with FPN, with longer 3x schedule. In the Tab.~\ref{tab:Abl:larger schedule}, SharpContour consistently improves the performance of baseline model. 

\begin{table}[t]
\resizebox{0.45\textwidth}{!}{
\begin{tabular}{c|c|cc}
\hline
            & backbone       &AP       & AP*           \\ \hline
Mask R-CNN                  & R50-FPN  & 37.2     & 39.5          \\
Mask R-CNN + SharpContour   & R50-FPN  & \textbf{39.3(+2.1)}     & \textbf{43.1(+3.6)}          \\
\hline
Mask R-CNN.                 & R101-FPN & 38.6     & 41.4         \\ 
Mask R-CNN + SharpContour   & R101-FPN & \textbf{40.8(+2.2)}    & \textbf{45.2(+3.8)}          \\
\hline
Mask R-CNN.                 & X101-FPN & 39.5     & 42.1       \\ 
Mask R-CNN + SharpContour   & X101-FPN & \textbf{41.8(+2.3)}   & \textbf{46.0(+3.9)}          \\\hline
\end{tabular}}
\centering
\vspace{-8pt}
\caption{\textbf{Larger models and longer 3x schedule}. Even with a stronger backbone (ResNet-101) and a longer training schedule (3x), a consistent improvement brought by SharpContour is achieved as well. AP* is COCO mask AP evaluated using the higher-quality LVIS annotations. }
\vspace{-10pt}
\label{tab:Abl:larger schedule}
\end{table}

\begin{table}[t]
\resizebox{0.38\textwidth}{!}{
\begin{tabular}{cc|ccc|ccc}
\hline
$\lambda$ & $AP_{val}$       & M  & $AP_{val}$       & FPS           & N   & $AP_{val}$      & FPS           \\ \hline
0.0015 & 37.1         & 5  & 36.8         & 15.4          & 128 & \textbf{37.5} & \textbf{15.0} \\
0.003  & \textbf{37.5} & 10 & \textbf{37.5} & \textbf{15.0} & 256 & 37.8         & 14.6  \\
0.006  & 36.6          & 15 & 37.2           & 14.7          & 348 & 38.0            & 14.2         \\
       &               & 20 & 37.1          & 14.3          & 512 & 38.0          & 13.7         \\ \hline
\end{tabular}}
\centering
\vspace{-8pt}
\caption{\textbf{$\mathrm{AP}_{val}$ and FPS for different parameters of sampling strategy on COCO val2017 datasets}. The parameters selected achieve the balance of accuracy and efficiency. }
\label{tab:Abl:SpinyContour}
\vspace{-12pt}
\end{table}

\begin{table}[t]
\centering
\resizebox{0.3\textwidth}{!}{
\begin{tabular}{c|cccc}
\hline
Combined Method      & n=1  & n=2  & n=3  & n=4  \\ \hline
DANCE                & 35.2 & 35.7 & 36.1 & 36.2 \\
Mask R-CNN           & 36.6 & 37.1 & 37.5 & 37.7     \\
CondInst             & 36.9 & 37.4 & 37.9 & 38.0     \\ \hline
\end{tabular}
}
\centering
\vspace{-8pt}
\caption{\textbf{Results of different iteration numbers of Evolution Process.}
    It can be seen that the Iterative Evolution can consistently boost the performance, as well as preserve stability on the COCO val2017 datasets. As the number of iterations increases, the improved performance of each iteration continues to decrease, indicating the convergence of contour. }
    \label{tab:Abl:Combination}
\vspace{-18pt}
\end{table}

\noindent \textbf{Different parameters of Contour Evolution process}
There are three main parameters in the design of the points sampling strategy, which are the deformation ratio $\lambda$, the number of sampling points $M$ and the resolution of the polygon $N$. In final version, we set $\lambda =0.003$, $M=10$ and $N=128$. In this ablation study, we conduct control experiments for each one of these parameters (preserving the other two parameters the same with the final version) and the results of three evolution iterations are in Tab.~\ref{tab:Abl:SpinyContour}. 

For $\lambda$, a) a larger deformation ratio causes performance degradation with more iterations, which is due to the fast deformation; b) a smaller deformation ratio requires more iterations for convergence. From these two experiments, we can see that our final choice achieves the best performance.

For $M$, a) a larger number of points will increase the inference time; b) a smaller number of points will decrease the running time but require more iterations for better performance; c) if the number of points becomes very large, the performance will decrease as more iterations are performed. This is because the polygon deforms too fast so that errors tend to appear in the result.

For $N$, it shows that denser sampling on the polygon contributes to minor performance improvements, while greatly increasing the inference time. We use 128-resolution in the final model for a better trade-off between accuracy and efficiency.

\noindent \textbf{Different number of Stacked Contour Evolution Process}
To comprehensively evaluate the effectiveness of the SharpContour, we explore the performance of different iterations of the Contour Evolution Process. The results are reported in Table.~\ref{tab:Abl:Combination}. As is shown, if stacking the contour evolution process of SharpContour more times, a stable enhancement of accuracy is validated for our methods. 

\begin{table}[h]
\centering
\vspace{-10pt}
\resizebox{0.3\textwidth}{!}{
\begin{tabular}{c|ccc}
\hline
   & Reg.1      & Reg. 2     & SharpContour        \\ \hline
AP & 36.1(+0.9) & 36.2(+1.0) & \textbf{37.5(+2.3)} \\ \hline
\end{tabular}
}
\centering
\vspace{-8pt}
\caption{\textbf{Results of Regression-based model}
    Both regression-based designs can only slightly enhance the AP. }
    \label{tab:Abl:Reg}
\vspace{-12pt}
\end{table}

\noindent \textbf{Regression-based Baselines} 
We have explored two regression-based designs before adopting current discretization design: 1) regressing offset vectors (\textbf{Reg.1}); 2)regressing the distance along the normal direction (\textbf{Reg. 2}), where the proposed instance-aware feature is also used. We adopted the Mask-RCNN as the basic framework and carried out experiments on COCO datasets. As is shown in Tab.~\ref{tab:Abl:Reg}, both regression-based designs can only slightly enhance the AP comparing with the SharpContour. 

\section{Conclusion}
We propose a novel contour-based refinement approach called SharpContour to improve the boundary quality of instance segmentation. The existing mask-based refinement methods are lack of efficiency or generality and contour-based approaches prone to generate over-smoothed contours around sharp corners. We address all of their limitations by designing a new contour evolution method and an Instance-aware Points Classifier. In contrast to the previous approaches, our approach shows superior accuracy in a considerable efficient fashion and can refine both mask-based and contour-based methods. We extensively evaluate the qualitative and quantitative performance of SharpContour on two large-scale public benchmarks for instance segmentation. It achieves significant improvement on all benchmarks when incorporated with different models. 


\vspace{-18pt}
\paragraph{Acknowledgement}
The work was supported in part by the Basic Research Project No. HZQB-KCZYZ-2021067 of Hetao Shenzhen-HK S$\&$T Cooperation Zone, National Key R$\&$D Program of China with grant No. 2018YFB1800800, by Shenzhen Outstanding Talents Training Fund 202002, and by Guangdong Research Projects No. 2017ZT07X152 and No. 2019CX01X104. This work was also supported by NSFC-62172348, 61902334, Shenzhen General Project (JCYJ20190814112007258) and Shenzhen Sustainable Development Project(KCXFZ20201221173008022). 

{\small
\bibliographystyle{ieee_fullname}
\bibliography{egbib}
}

\end{document}